\documentclass[sn-mathphys,Numbered]{sn-jnl}

\usepackage{graphicx}%
\usepackage{multirow}%
\usepackage{amsmath,amssymb,amsfonts}%
\usepackage{amsthm}%
\usepackage{mathrsfs}%
\usepackage[title]{appendix}%
\usepackage{xcolor}%
\usepackage{textcomp}%
\usepackage{manyfoot}%
\usepackage{booktabs}%
\usepackage{algorithm}%
\usepackage{algorithmicx}%
\usepackage{algpseudocode}%
\usepackage{listings}%

\usepackage{multirow}
\usepackage{natbib}
\usepackage{tabularx}
\usepackage{bm}
\usepackage{float}
\usepackage[caption=false]{subfig}

\graphicspath{{./pdffigure/}} 

\theoremstyle{thmstyleone}%
\theoremstyle{thmstyletwo}%

\theoremstyle{thmstylethree}%

\begin{document}

\title[Article Title]{Dual-Context Aggregation for Universal Image Matting}


\author*{\fnm{Qinglin} \sur{Liu}}

\author*{\fnm{Xiaoqian} \sur{Lv}}

\author{\fnm{Wei} \sur{Yu}}

\author{\fnm{Changyong} \sur{Guo}}

\author{\fnm{Shengping} \sur{Zhang}}



\abstract{Natural image matting aims to estimate the alpha matte of the foreground from a given image. 
Various approaches have been explored to address this problem, such as interactive matting methods that use guidance such as click or trimap, and automatic matting methods tailored to specific objects. 
However, existing matting methods are designed for specific objects or guidance, neglecting the common requirement of aggregating global and local contexts in image matting.
As a result, these methods often encounter challenges in accurately identifying the foreground and generating precise boundaries, which limits their effectiveness in unforeseen scenarios.
In this paper, we propose a simple and universal matting framework, named Dual-Context Aggregation Matting (DCAM), which enables robust image matting with arbitrary guidance or without guidance. 
Specifically, DCAM first adopts a semantic backbone network to extract low-level features and context features from the input image and guidance. 
Then, we introduce a dual-context aggregation network that incorporates global object aggregators and local appearance aggregators to iteratively refine the extracted context features.
By performing both global contour segmentation and local boundary refinement, DCAM exhibits robustness to diverse types of guidance and objects.
Finally,  we adopt a matting decoder network to fuse the low-level features and the refined context features for alpha matte estimation.
Experimental results on five matting datasets demonstrate that the proposed DCAM outperforms state-of-the-art matting methods in both automatic matting and interactive matting tasks, which highlights the strong universality and high performance of DCAM.
The source code is available at \url{https://github.com/Windaway/DCAM}.
}




\maketitle

\section{Introduction}\label{sec1}
\label{sec:intro}
Natural image matting is a fundamental technology in the field of computer vision and computer graphics that aims to estimate the alpha matte of the foreground in a given image. 
This technology finds wide-ranging applications in multimedia domains, including image editing~\cite{2009Sketch2Photo,2017Robust}, live streaming~\cite{gastal2010shared,2015Integrated,lin2021real}, and augmented reality~\cite{zongker1999environment,li2022matting}, with significant commercial value.  
Therefore, it has been extensively studied by numerous researchers. 
Formally, a given image $\bm{I}$ can be represented as a combination of the foreground $\bm{F}$ and background $\bm{B}$ as
\begin{equation}
\label{eq:pre}
{{I}}^i = {\alpha}^i {F}^i + (1-{ \alpha}^i ){B}^i 
\end{equation}
where ${\alpha}^i$ is the alpha matte of the foreground at pixel $i$. 
Since only the image $\bm{I}$ is known in Equation~\ref{eq:pre}, directly predicting alpha matte from arbitrary images is extremely challenging.
Therefore, researchers have focused on designing automatic matting methods for specific objects or interactive matting methods using guidance like click or trimap, and significant progress has been achieved.

Early researchers have focused on using manually designed rules to develop sampling-based or propagation-based matting methods. 
Sampling-based methods utilize statistical information of texture or color within an image to estimate the alpha matte, while propagation-based methods use the smoothness of foreground and background color in local regions to propagate alpha matte from known to unknown regions.
However, due to the simplistic assumptions made by these traditional matting methods, they struggle to handle complex color or texture distributions in real-world scenarios, resulting in suboptimal performance.
In recent years, researchers have  shifted towards the use of neural networks to address the matting problem.
These methods involve  building neural networks and training  them on matting datasets to estimate the alpha matte.
The learning capability of neural networks allows them to capture the contextual information of color, texture, object structure, and guidance for alpha matte prediction.
As the matting datasets contain more complex priors than manually designed rules, learning-based methods achieve significant performance improvement over traditional methods.

Learning-based image matting methods have made significant progress on various matting benchmarks. 
However, they are usually designed for specific guidance or objects, and their performance deteriorates when applied to other guidance or objects. 
Thus, designing a specific neural network architecture and performing finetuning, which requires expert knowledge and is time-consuming, is often necessary for new matting tasks. 
The reason for this is that existing learning-based matting methods do not consider global and local context aggregation in the matting network.
Specifically, global context aggregation helps the network identify the object contours under coarse or no guidance, thereby enhancing the universality of the matting network to handle various types of guidance.
Local context aggregation assists the network in identifying object boundaries, thereby improving the matting accuracy.
Combining global-local context aggregation can enhance the universality of the matting networks while improving the matting performance.

In this paper, we propose a simple and universal matting framework, named Dual-Context Aggregation Matting (DCAM), which can perform robust image matting with arbitrary guidance or without guidance. 
To this end, we introduce a dual-context aggregation module into the basic encoder-decoder network to perform global and local context aggregation. 
Specifically, DCAM first uses a semantic backbone network to extract low-level features and  context features from the input image and guidance. 
Then, we introduce a dual-context aggregation network that incorporates global object aggregators and local appearance aggregators to iteratively refine the extracted context features.
Notably, the global object aggregator utilizes semantic-instance attention to perform global contour refinement,  while the local appearance aggregator adopts a hybrid transformer structure that utilizes both low-frequency and high-frequency context to perform local segmentation refinement.
These designs greatly improve the robustness of DCAM to diverse guidance and objects.
Finally, we adopt a matting decoder network to fuse low-level features and refined context features for alpha matte estimation.
Experimental results on five image matting datasets demonstrate that DCAM outperforms state-of-the-art matting methods in both automatic and interactive matting tasks, which indicates the strong universality and  high performance of DCAM.

In summary, the contributions of this paper are threefold:

\begin{itemize}
  \item We propose a simple and universal matting framework, named Dual-Context Aggregation Matting (DCAM), which enables robust image matting with arbitrary guidance or without guidance. 
  \item  We propose a dual-context aggregation network that includes global object aggregators and local appearance aggregators to iteratively refine the extracted context features, which improve the robustness of DCAM to diverse guidance and objects.

  \item Extensive experimental results on five matting datasets, namely HIM-100K, Adobe Composition-1K, Distinctions-646, P3M, and PPM-100, demonstrate that the proposed DCAM outperforms state-of-the-art matting methods in both automatic and interactive matting tasks.
\end{itemize}

The paper is organized as follows: In Section II, we provide an overview of traditional sampling-based or propagation-based matting methods and recent learning-based matting methods relevant to our work. 
In Section III, we present a detailed introduction to our Dual-Context Aggregation Matting (DCAM).
In Section IV, we describe the training details of DCAM, including the optimizer and hyperparameters. 
We also discuss  the datasets used for evaluation and provide a comprehensive  comparison of DCAM with state-of-the-art matting methods. 
Finally, in Section V, we conclude our work and discuss future directions for matting research. 

\section{Related Works}
\subsection{Traditional matting methods.}
Traditional image matting methods can be divided into two categories: sampling-based methods and propagation-based methods. Sampling-based methods~\cite{berman1998method,ruzon2000alpha,wang2007optimized,gastal2010shared,he2011a,shahrian2013improving} use statistical information of color and texture in the image to sample candidate foreground and background colors for unknown pixels, which are used to estimate the alpha matte. These methods improve the efficiency or accuracy by optimizing the sampling process.
Berman \emph{et al.}~\cite{berman1998method} first propose to estimate the alpha matte by sampling foreground and background colors in the known regions around unknown pixels. 
Bayesian Matting~\cite{chuang2001a} uses a Gaussian distribution to model the color and location information of foreground and background to help estimate alpha matte.
Robust Matting~\cite{wang2007optimized} uses the Random Walk with a matting energy function to sample foreground and background colors for robust alpha matte estimation. 
Shared Matting~\cite{gastal2010shared} utilizes the assumption that foreground or background colors are similar in a local region to accelerate inference. 
Global Matting~\cite{he2011a} proposes to sample the pixels in all known regions to avoid information loss. 
Shahrian \emph{et al.}~\cite{shahrian2013improving} use an object function to estimate the color distribution at known regions for sampling, which improves the matting accuracy.

Propagation-based matting methods~\cite{sun2004poisson,grady2005random,levin2008a,levin2008spectral,he2010fast,chen2013knn,li2013motion,aksoy2017designing} assume that foreground and background colors are continuous within local regions and propagate alpha matte from known regions to unknown regions. These methods optimize the propagation function to improve efficiency or accuracy.
Poisson Matting~\cite{sun2004poisson} solves the Poisson equation using boundary information from the trimap.
Random Walk~\cite{grady2005random} approximates the alpha matte with the probability that a random walker leaving an unknown pixel reaches a foreground pixel before reaching a background pixel.
Close-form matting~\cite{levin2008a} uses the color-line model to provide a closed-form solution for matting.
He \emph{et al.}~\cite{he2010fast}  enlarge the kernel of Laplace matrix to accelerate the inference
KNN matting~\cite{chen2013knn} improves the similarity matrix and objective function using the nonlocal principle.

\subsection{Learning-based matting methods.}
Learning-based image matting methods utilize neural networks to predict alpha matte and can be divided into interactive matting methods and automatic matting methods. Interactive matting methods~\cite{xu2017deep,tang2019learning,cai2019disentangled,lu2019indices,li2020natural,forte2020fbamatting,2020Context,yu2020high,yu2020mask,wang2021ImprovingDeepImageMatting,Liu_2021_ICCV,sun2021sim,dai2022boosting,park2022matteformer} uses additional information such as trimap or mask to assist the network in predicting the alpha matte.
DIM~\cite{xu2017deep} proposes the first end-to-end matting network with a large-scale image matting dataset.
SampleNet~\cite{tang2019learning} first predicts the foreground and background and then uses these predictions to help estimate the alpha matte. 
IndexNet~\cite{lu2019indices} employs the indice information to help estimate alpha mattes of high gradient accuracy and visual quality.
GCAMatting~\cite{li2020natural} uses a Guided Contextual Attention module to predict the alpha matte of semi-transparent objects with contextual information.
HDMatt~\cite{yu2020high}  estimates alpha mattes through patch-based inference with a Cross-Patch Contextual module. 
TIMI~\cite{Liu_2021_ICCV} proposes to mine the relationship between global and local features to improve predictions. 
SIM~\cite{sun2021sim} incorporates semantic segmentation into the matting network  to improve predictions in complex scenes. 
MatteFormer~\cite{park2022matteformer} uses Swin~\cite{liu2021Swin} transformer to extract long-range contexts, achieving good performance in complex scenes.

Automatic matting methods~\cite{chen2018semantic,Zhang2019A,qiao2020attention,ke2020is,Yu_2021_ICCV,2021Privacy,li2022matting,liu2020boosting} automatically predict the alpha matte of the foreground object (usually human) in the image.
SHM~\cite{chen2018semantic} proposes the usage of neural networks for estimating trimaps followed by human matting.
LFM~\cite{Zhang2019A} proposes to use a dual-branch network to predict the segmentation of the foreground and the background, and then fuses the predictions to generate the final alpha matte.
Srivastava \emph{et al.}~\cite{srivastava2022alpha} use an encoder-decoder network to directly predict the alpha matte.
BSHM~\cite{liu2020boosting} employs three networks for segmentation, refinement, and matting processes, which enhances the estimated alpha mattes.
HAttMatting~\cite{qiao2020attention}  proposes the use of hierarchical attention modules to learn multi-scale context features, thereby improving the alpha matte estimation.
CasDGR~\cite{Yu_2021_ICCV} adopts deformable graph refinement for refining the features extracted by residual U-Net, and then estimates the alpha matte.
MODNet~\cite{ke2020is} proposes a self-supervised learning strategy for improving the generative ability of matting methods with unlabeled real-world data.
P3M~\cite{2021Privacy} designs a two-branch decoder to estimate the segmentation and boundary alpha matte, which improves the robustness of the method.
GFM~\cite{li2022matting} utilizes a novel composition pipeline to improve the performance of matting networks in real-world scenes.

\begin{figure*}[!t]
\resizebox{\linewidth}{!} {
  \includegraphics[width=0.98\linewidth]{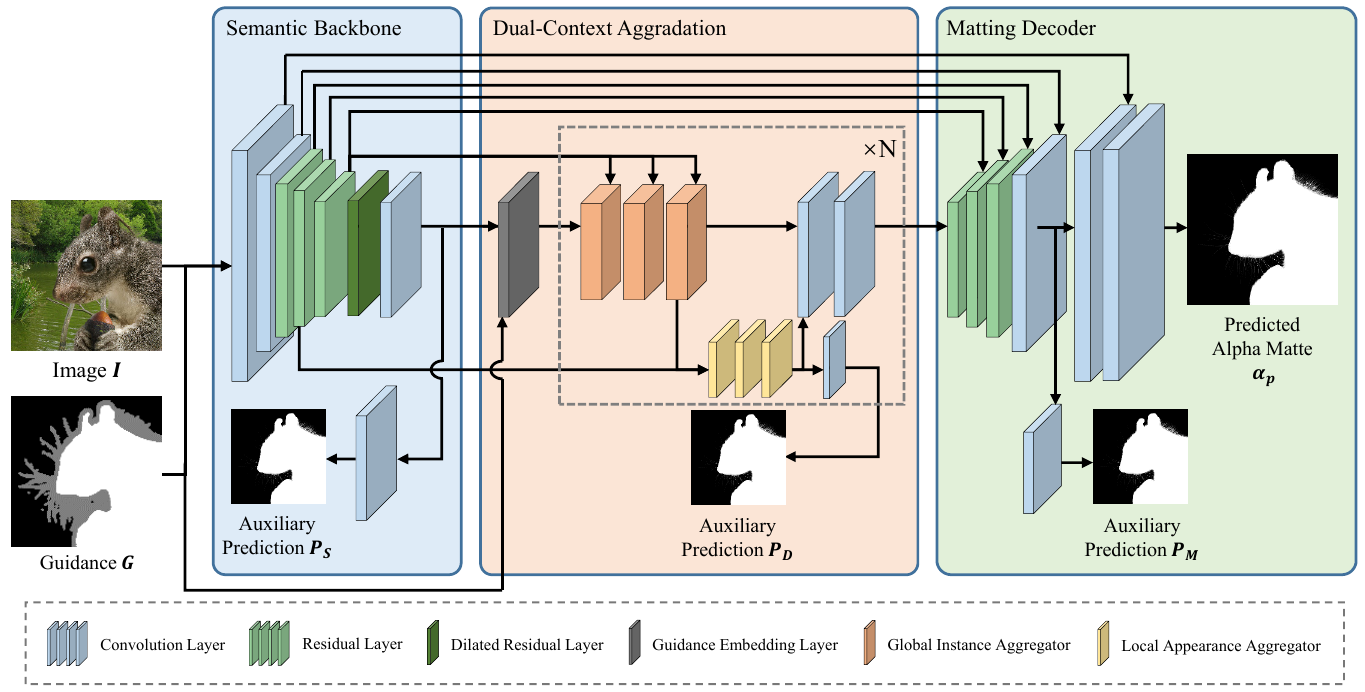}
    }
  \caption{Overview of the Dual-Context Aggregation Matting (DCAM) framework. A semantic backbone network first extracts low-level features and context features from the  input image and guidance. 
    Then, a dual-context aggregation network iteratively performs global object aggregation and local appearance aggregation to refine the extracted context features. 
    Finally, a matting decoder network fuses the low-level features with the refined context features to predict the alpha matte.}
\vspace{-3pt}
  \label{fig:overview}
\end{figure*}

\section{Our Method}
The proposed DCAM framework adopts a U-Net style architecture with a dual-context aggregation design  to improve the predicted alpha mattes through global-local context aggregation.
As shown in Figure~\ref{fig:overview}, a semantic backbone network first extracts low-level features and context features from the  input image and the corresponding guidance. 
Then, a dual-context aggregation network is employed to iteratively perform global object aggregation and local appearance aggregation to refine the extracted context features. 
Finally, a matting decoder network fuses the low-level features with the refined context features to predict the alpha matte.
\subsection{Semantic Backbone Network}
The semantic backbone network aims to extract low-level features and context features from the input image $\bm{I}$ and guidance $\bm{G}$, which are utilized for subsequent dual-context aggregation and alpha matte prediction.
To achieve this, we employ a convolutional hierarchical encoder structure with auxiliary prediction tasks.
Specifically, we concatenate the image $\bm{I}$ and optional guidance $\bm{G}$ together, where $\bm{G}$ can be either trimaps or clicks for interactive matting, or no guidance for automatic matting.
To extract rich low-level features for alpha matte estimation, we construct a deep stem using $3 \times 3$ convolution layers, which preserves more image details compared to the patch embedding layers of transformer-based networks~\cite{liu2021Swin,park2022matteformer}.
Next, we adopt the residual blocks of ResNet-50~\cite{he2016deep} to extract context features.
Given the small batch size in training matting networks, we utilize group normalization instead of batch normalization to stabilize network training. 
Furthermore, we replace the fourth residual block of ResNet-50 with an atrous convolutional residual block, which removes the downsampling operations and preserves spatial information in the context features.
Finally, we employ $1 \times 1$ convolution layers to reduce the dimensionality of the extracted features and obtain compact context features $\bm{F}_c$. 
These features are used to generate auxiliary predictions $\bm{P}_S$ using $3 \times 3$ convolution layers.
Depending on the guidance, the auxiliary prediction is either trimaps for the click guidance and no guidance, or alpha mattes for the trimap guidance.

\subsection{Dual-Context Aggregation Network}
The dual-context aggregation network aims to aggregate and refine the context features extracted by the semantic backbone network to help the subsequent alpha matte prediction.
To this end, we employ a guidance embedding layer to project the guidance to features and then integrate them into the context features. 
Then, we adopt the global object aggregators and local appearance aggregators to aggregate the guidance-enhanced context features.
Finally, we cascade the aggregators twice and introduce auxiliary predictions to iteratively refine the context features.

\begin{figure*}[!t]
  \centering
  \subfloat[Global Object Aggregator]{
  \label{fig:glr}
            \includegraphics[width=0.99\linewidth]{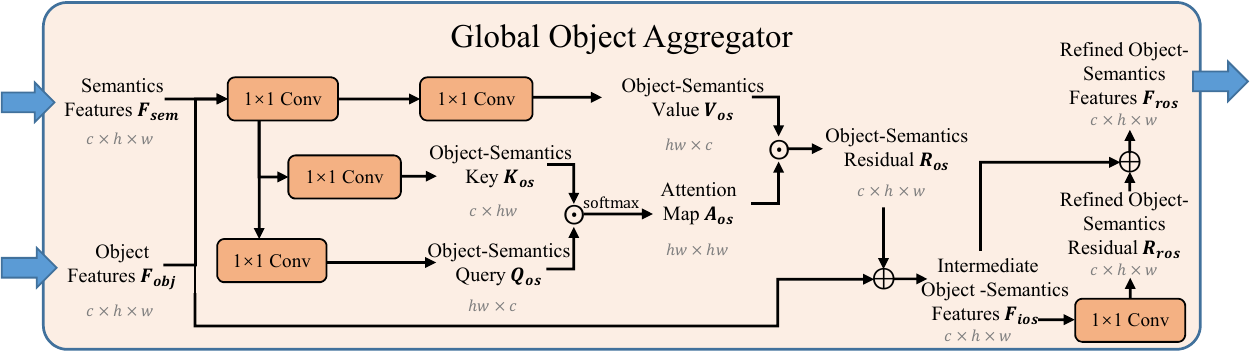}
		}
		
  \subfloat[Local Appearance Aggregator]{\label{fig:loc}
			\includegraphics[width=.99\linewidth]{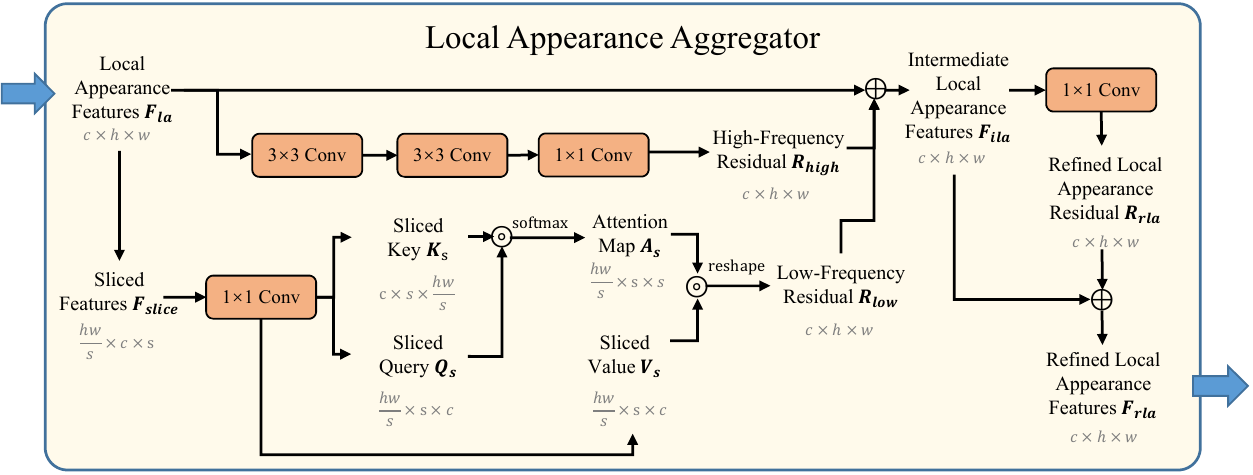}
		}
  \caption{Structures of the global object aggregator and the local appearance aggregator. The  global object aggregator utilizes semantic-object attention to perform global contour refinement,  while the local appearance aggregator adopts a hybrid transformer structure that utilizes both low-frequency and high-frequency context to perform local segmentation refinement.}
  \label{fig:ghs}
\end{figure*}

\noindent \textbf{Guidance Embedding Layer.}
The guidance information is critical for interactive matting. 
However, the semantic backbone network usually dilutes the impacts of guidance during feature extraction, which may hinder the context aggregation based on the guidance.
To address this issue, we propose a simple yet effective guidance embedding layer to enhance the guidance for the foreground objects. 
Given the guidance $\bm{G}$, we first use a simple fully connected layer to generate the guidance features $\bm{F}_g$ as $\bm{F}_g={\rm MLP}(G)$, where ${\rm MLP}$ denotes a multi-layer perceptron. 
We then integrate $\bm{F}_g$ and the context features $\bm{F}_c$ through the element-wise addition to obtain the object features $\bm{F}_{obj}$ as $\bm{F}_{obj} = \bm{F}_c + \bm{F}_g$.

\noindent \textbf{Global Object Aggregator.} 
Estimating the contour of the objects is crucial for predicting alpha matte in cases where the guidance information is coarse.
Recent semantic segmentation methods have demonstrated  that utilizing the feature affinity of objects in the image helps estimate segmentation~\cite{fu2020scene,YuanCW19,WangSCJDZLMTWLX19,SunXLW19}.
Inspired by these findings, we propose to use attention to aggregate guidance-enhanced context features.
Furthermore,  we use semantic features from shallower layers in the semantic backbone network since they contain more detailed information and helps distinguish differences between objects, thus improving the aggregation.

The structure of the global object aggregator is shown in Figure~\ref{fig:glr}. Specifically, we take the object features $\bm{F}_{obj}$ from the guidance embedding layer and the semantic features $\bm{F}_s$ from the third residual block of the semantic backbone network as inputs. We first use a $1 \times 1$ convolution layer to perform residual fusion of $\bm{F}_{obj}$ and $\bm{F}_s$ and obtain the object-semantic feature $\bm{F}_{os}$ as
\begin{equation}
\bm{F}_{os} =\bm{F}_{obj}+ \text{Conv}(\text{Concat} (\bm{F}_{obj} , \bm{F}_s))
\end{equation}
where $\text{Conv}(\cdot)$ denotes the $1 \times 1$ convolution layer. $\text{Concat}(\cdot , \cdot)$ denotes the concatenate operation.
Then, we use $\bm{F}_{os}$ to generate the corresponding object-semantics query features $\bm{Q}_{os}$, object-semantics key features $\bm{K}_{os}$, and object-semantics value features $\bm{V}_{os}$ using $1 \times 1$ convolution layers.
Next, we compute the attention map $\bm{A}_{os}$ using $\bm{Q}_{os}$ and $\bm{K}_{os}$ as
\begin{equation}
\bm{A}_{os} =\text{Softmax}(\bm{Q}_{os}\bm{K}_{os}^\top)
\end{equation}
where $\text{Softmax}(\cdot)$ denotes the softmax operation.
Subsequently, we use the attention map $\bm{A}_{os}$ to perform a weighted fusion with the object-semantics value features $\bm{V}_{os}$ to obtain the object-semantic residual $\bm{R}_{os}$ as
\begin{equation}
\bm{R}_{os} =\bm{A}_{os} \bm{V}_{os}
\end{equation}
Afterward, we perform element-wise addition fusion of the object features and object-semantics residual $\bm{R}_{os}$ to obtain the intermediate object feature $\bm{F}_{iobj}$ as $\bm{F}_{iobj} =\bm{F}_{obj} + \bm{R}_{os}$.
We use a $1 \times 1$ convolution layer to obtain the refined object-semantics residual $\bm{R}_{ros}$.
Finally, we perform element-wise addition fusion of the intermediate object features $\bm{F}_{iobj}$ and refined object-semantics residual $\bm{R}_{ros}$ to obtain the refined object-semantics features $\bm{F}_{robj}$ as $\bm{F}_{robj} =\bm{F}_{iobj} + \bm{R}_{ros}$.

\noindent \textbf{Local Appearance Aggregator.}
Accurate boundary segmentation is crucial for predicting the alpha mattes in the boundary. 
Aggregating local context features with appearance information helps to distinguish boundary segmentation. 
Moreover, previous studies have shown that both low-frequency and high-frequency information are important for segmentation tasks~\cite{dong2023afformer}. 
Therefore, we propose to adopt a hybrid transformer structure for local context aggregation.

The structure of the local appearance aggregator is depicted in Figure~\ref{fig:loc}.
Specifically, we use a $1\times 1$ convolution to fuse the features from the second residual block of the semantic backbone network and the upsampled refined object-semantics features as the input local appearance features $\bm{F}_{la}$.
We then employ parallel CNN and transformer paths to respectively aggregate high-frequency and low-frequency information. 
In the CNN path, we generate the high-frequency residual $\bm{R}_{high}$ using two $3\times 3$ convolutions and a $1\times 1$ convolution sequentially.
In the transformer path, we first slice the local appearance features $\bm{F}_{la}$ into several sliced features $\bm{F}_{slice}$ of size $s\times s$. We then generate Query, Key, and Value features $\bm{Q}_{slice}$, $\bm{K}_{slice}$, and $\bm{V}_{slice}$ for each window feature using $1\times 1$ convolution layers. 
Next, we compute the attention map $\bm{A}_{slice}$ for each window feature slice by calculating the Query-Value similarity with the Key features as
\begin{equation}
\bm{A}_{slice} =\text{Softmax}(\bm{Q}_{slice}\bm{K}_{slice}^\top)
\end{equation}
We then use the attention map $\bm{A}_{slice}$ to weight the fusion of the Value features $\bm{V}_{slice}$ and reshape it to obtain the low-frequency residual $\bm{R}_{low}$ as
\begin{equation}
\bm{R}_{low} ={\rm Reshape} (\bm{A}_{slice} \bm{V}_{slice})
\end{equation}
where ${\rm Reshape}$ denotes the reshape operation. We then perform element-wise addition of the intermediate local appearance features $\bm{F}_{ila}$ and the refined local-appearance residual $\bm{R}_{rla}$ to obtain the refined local-appearance features $\bm{F}_{rla}$ as $\bm{F}_{rla} =\bm{F}_{ila} + \bm{R}_{rla}$, where $\bm{F}_{ila}=\bm{F}_{la}+\bm{R}_{high}+\bm{R}_{low}$, and we use a $1\times 1$ convolution layer to obtain the refined local-appearance residual $\bm{R}_{rla}$.

\subsection{Matting Decoder Network}
The matting decoder network aims to fuse the low-level features from the semantic backbone network and the refined context features from the dual-context aggregation network to predict the alpha mattes. 
To achieve this, we adopt a hierarchical fully convolutional network to progressively upsample and refine the context features. 
In addition, we introduce auxiliary predictions to  facilitate network training or assist in foreground-background classification.
Specifically, we first concatenate the refined context features from the dual-context aggregation network with the semantic features from the third block of the semantic backbone network.
We then employ a $1 \times 1$ convolution to compress the feature dimension and residual layers to refine the context features. 
We use the same method to refine and upsample the context features with the appearance features from the second and first blocks of the semantic backbone network. 
Next, we concatenate the context features with the low-level features from the stem of the semantic backbone network and use $3 \times 3$ convolution to fuse them to obtain the matte features. 
We use the matte features to generate auxiliary predictions $\bm{P}_M$, which are trimaps when there is no guidance or click guidance, and alpha mattes when there is trimap guidance.
Finally, we upsample the matte features and concatenate them with the input image $\bm{I}$, and use $3 \times 3$ convolution layers to predict the final alpha matte $\bm{\alpha}_p$.

\subsection{Loss Function}
To train the proposed DCAM framework, we design loss functions for the semantic backbone network, dual-context aggregation network, and matting decoder network. 
Specifically, we adopt different types of auxiliary outputs for different types of guidance. 
For the case of coarse guidance like click or no guidance, all auxiliary outputs are the trimap of the image. 
In this case, the semantic backbone loss $\mathcal{L}_S$ is defined as
\begin{equation}
\mathcal{L}_S = {\rm FocalCE} (\bm{P}_S,\bm{T}_{gt})
\end{equation}
where $ {\rm FocalCE}$ is the focal crossentropy function. $\bm{P}_S$ is the auxiliary prediction of the semantic backbone, and $\bm{T}_{gt}$ is the ground truth trimap.
The dual-context aggregation loss $\mathcal{L}_D$ is defined as 
\begin{equation}
\mathcal{L}_D = {\rm FocalCE} (\bm{P}_D,\bm{T}_{gt})
\end{equation}
where $\bm{P}_D$ is the auxiliary prediction of the dual-context aggregation.
The matting decoder loss $\mathcal{L}_M$ is defined as 
\begin{equation}
\mathcal{L}_M ={\rm FocalCE} (\bm{P}_M,\bm{T}_{gt})+ \frac{1}{|{T}^U|}\sum_{{i\in{T}_U}}{\sqrt{(\alpha_p^i-\alpha^i_{gt})^2+\epsilon^2}}\\
\end{equation}
where $\bm{P}_M$ is the auxiliary prediction of the matting decoder. $\bm{T}_U$ is the set of all unknown pixels in the trimap $\bm{T}_{gt}$. ${\alpha}_p^i$ and ${\alpha}_{gt}^i$ are the predicted alpha matte and ground truth alpha matte at pixel $i$, respectively. 
$\epsilon$ is the penalty coefficient. 

For the case of trimap guidance, all auxiliary outputs are the alpha matte of the foreground. In this case, the semantic backbone loss $L_S$ is defined as
\begin{equation}
\mathcal{L}_S = \frac{1}{|{T}^U|}\sum_{{i\in\bm{T}_U}}{\sqrt{(P_S^i-\alpha^i_{gt})^2+\epsilon^2}}\\
\end{equation}
where $\bm{T}_U$ is the set of all unknown pixels in the trimap $\bm{T}_{gt}$. ${P}_S^i$ and $\bm{\alpha}_{gt}^i$ are the auxiliary prediction of semantic backbone and ground truth alpha matte at pixel $i$. 
The dual-context aggregation loss $\mathcal{L}_D$ is defined as 
\begin{equation}
\mathcal{L}_D = \frac{1}{|{T}^U|}\sum_{{i\in\bm{T}_U}}{\sqrt{(P_D^i-\alpha^i_{gt})^2+\epsilon^2}}\\
\end{equation}
where ${P}_D^i$ is the auxiliary prediction of the dual-context aggregation at pixel $i$.
The matting decoder loss $\mathcal{L}_M$ is defined as 
\begin{equation}
\begin{aligned}
    \mathcal{L}_M &= \frac{1}{|{T}^U|}\sum_{{i\in\bm{T}_U}}{\sqrt{(P_M^i-\alpha^i_{gt})^2+\epsilon^2}}\\
    &+\frac{1}{|{T}^U|}\sum_{{i\in\bm{T}_U}}{\sqrt{(\alpha_p^i-\alpha^i_{gt})^2+\epsilon^2}}\\
    &+\sum_{j}{2^{j}\|{\rm L}_j(\bm{\alpha}_p)-{\rm L}_j(\bm{\alpha}_{gt})\|_1}
\end{aligned}
\end{equation}
where ${P}_M^i$ is the auxiliary prediction of the matting decoder at pixel $i$.
$\bm{\alpha}_p$ is the predicted alpha matte. 
${\rm L}_j(\bm{\alpha}_p)$ and ${\rm L}_j(\bm{\alpha}_{gt})$ are the $j$-th level of the Laplacian pyramid representations of $\bm{\alpha}_p$ and $\bm{\alpha}_{gt}$, respectively.

The total loss of DCAM is defined as
\begin{equation}
\mathcal{L}= \mathcal{L}_S + \mathcal{L}_D + \mathcal{L}_M
\end{equation}

\section{Experiments}
In this section, we perform comprehensive experiments on five datasets to validate the effectiveness of the proposed DCAM framework. 
Firstly, we introduce the implementation details of DCAM and the datasets utilized for training and evaluation. 
Then, we compare DCAM with existing image matting methods on these datasets.
Finally, we conduct ablation studies to demonstrate the effectiveness of the proposed method.

\subsection{Datasets}
In this paper, we perform comprehensive experiments on the HIM-100K~\cite{10224299}, Adobe Composition-1K~\cite{he2016deep}, Distinctions-646~\cite{qiao2020attention}, P3M~\cite{2021Privacy}, and PPM-100~\cite{ke2020is} datasets.

\noindent \textbf{Human Instance Matting (HIM-100K)} is a instance-level human matting dataset, which comprises a training set and a testing set.
The training set contains 100,000 real-world or synthetic human group photos, while the testing set contains 1,500 real-world human group photos. 
Each image in both the training and testing sets includes corresponding alpha matte annotations for each human instance, which is utilized for evaluating the click-based interactive matting methods.

\noindent \textbf{Adobe Composition-1K} is a general object matting dataset that consists of a training set and a test set. 
The training set comprises 43,100 images synthesized from 431 foreground images, while the test set comprises 1,000 images synthesized from 50 foreground images.
This dataset is utilized for evaluating the trimap-based interactive matting methods and automatic matting methods.

\noindent \textbf{Distinctions-646} is also a general object matting dataset that comprises a training set and a test set. 
The training set comprises 59,600 images synthesized from 596 foreground images, while the test set comprises 1,000 images synthesized from 50 foreground images.
This dataset is utilized for evaluating the trimap-based interactive matting methods and automatic matting methods.

\noindent \textbf{Privacy-Preserving Portrait Matting (P3M)} is a privacy-preserving portrait matting dataset. The dataset includes a training set and two test sets. The training set consists of 9,421 portrait images of blurred faces. The P3M-500-P  test set includes 500 images of blurred faces, and the P3M-500-NP test set  includes 500 images of normal faces. This dataset is used to train and evaluate automatic matting methods.

\noindent \textbf{Photographic Portrait Matting (PPM-100)} is a portrait matting benchmark. 
The test set of PPM-100 consisting of 100 well-annotated portrait images.
Due to the diverse humans and backgrounds of the images, PPM-100 is a valuable benchmark for evaluating the generalization ability of  matting methods.

\subsection{Implementation Details}
The proposed DCAM is implemented with PyTorch~\cite{NEURIPS2019_9015} and trained on the training set of the HIM-100K, Adobe Composition-1K, Distinctions-646, and P3M datasets with four NVIDIA RTX 2080Ti GPUs. The code and model will be made available to the public.
Specifically, we utilize the Kaiming initializer~\cite{he2015delving} to initialize the network weights.
To accelerate the training, we initialize the encoder of DCAM with the weights of Resnet-50~\cite{he2016deep} pre-trained on ImageNet~\cite{deng2009imagenet}.
In addition, we set the coefficients in the network architecture and network loss function as $s=7$,  $j=4$, $\epsilon=1e-6$.
Then, we use Adam optimizer~\cite{kingma2014adam} with betas $(0.5,0.999)$ and weight decay of $1e-5$ to train the network. 
The initial learning rate for the optimizer is set to $1e-4$ and is decreased to $1e-7$ during training using the ``Cosine Annealing" scheduler.
To prevent the DCAM network from overfitting, we follow GCAMatting~\cite{li2020natural} and P3M~\cite{2021Privacy} to adopt data augmentation methods, such as random resizing, random color transformation, and random cropping to process the training data.
Finally, we train the DCAM network for 50 epochs using a total batch size of 8.

\subsection{Experimental Results}

\subsubsection{Results on HIM-100K}
\begin{figure*}[]
	\centering{
			\includegraphics[width=1\linewidth]{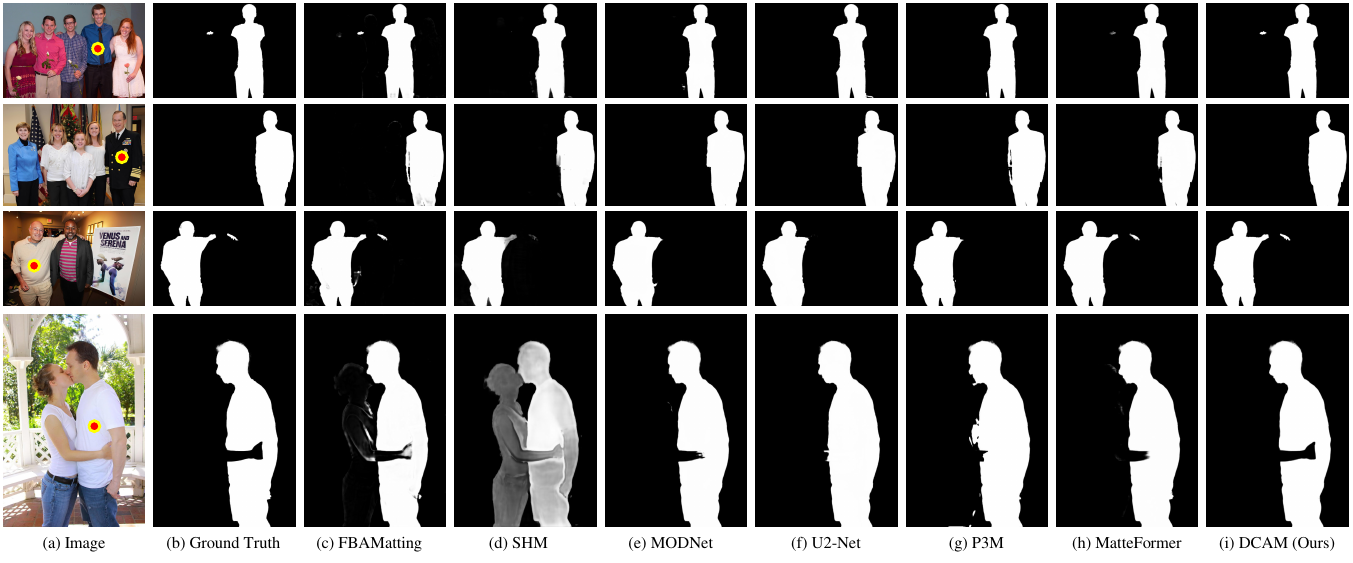}
	}
	\caption{Qualitative results on the HIM-100K dataset. The red dots denote the click guidance.}
	\label{exp_real}
\end{figure*}

To evaluate the performance of the proposed DCAM under click guidance, we compare it with existing interactive and automatic matting methods, including IndexNet~\cite{lu2019indices}, FBAMatting~\cite{forte2020fbamatting}, SHM~\cite{chen2018semantic}, U2Net~\cite{Qin_2020_PR}, MODNet~\cite{ke2020is}, P3M~\cite{2021Privacy}, and  MatteFormer~\cite{park2022matteformer}, on the HIM-100K dataset. 
The input layer of existing methods is modified to accept click guidance as input.
We train DCAM and the compared methods on the HIM-100K dataset, and follow MODNet to adopt MAD and MSE as evaluation metrics. The qualitative and quantitative results are summarized in Table~\ref{table_real_sel} and Figure~\ref{exp_real}, respectively.
Table~\ref{table_real_sel} shows that P3M performs better than existing interactive and automatic matting methods.
However, our DCAM significantly outperforms all existing methods by a large margin. 
Figure~\ref{exp_real} illustrates that our DCAM performs much better at removing interference from other humans or backgrounds and achieves better visual perception than existing matting methods.
The quantitative and qualitative results indicate that the proposed DCAM significantly outperforms existing matting methods under the click guidance.

\begin{table*}[!t]
	\caption{Quantitative results on the HIM-100K dataset. }
	\label{table_real_sel}
	\centering
	\begin{tabular}{l|cc}
    \toprule
	Method  & MSE & MAD \\
    \midrule
    IndexNet~\cite{lu2019indices} &0.12770 &0.15666\\
    FBAMatting~\cite{forte2020fbamatting} &0.00411 &0.00845\\  
	SHM~\cite{chen2018semantic}  &0.00399 &0.00997\\
	MODNet~\cite{ke2020is} &0.00557&	0.00756\\
	U2-Net~\cite{Qin_2020_PR} &0.00516	&0.00678\\ 
    P3M~\cite{2021Privacy} & 0.00337&	0.00473 \\
     MatteFormer~\cite{park2022matteformer} &{0.00386}&{0.00655}\\ 
	\midrule
	DCAM (Ours)  &\bf{0.00228}  & \bf{ 0.00342}\\
    \bottomrule
	\end{tabular}
\end{table*}

\subsubsection{Results on Adobe Composition-1K}
We evaluate the performance of DCAM on the trimap-based interactive matting and automatic matting tasks using the Adobe Composition-1K dataset.

\begin{figure*}[!t]
	\centering{
			\includegraphics[width=1\linewidth]{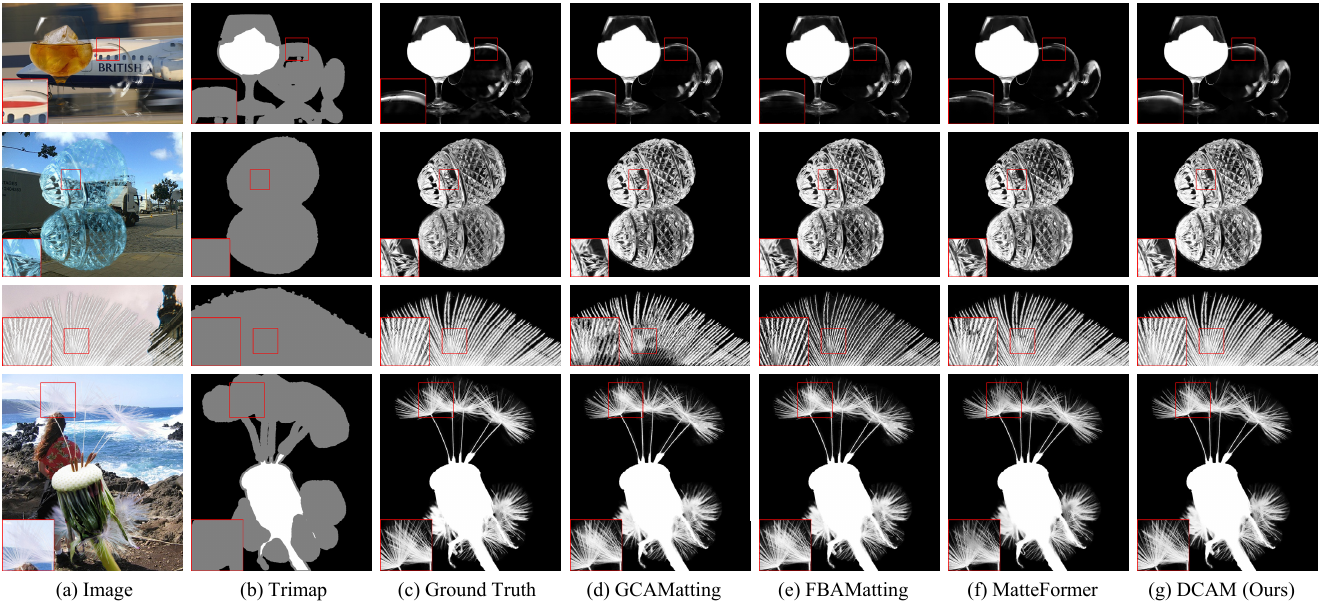}
	}
	\caption{Qualitative results on the Adobe Composition-1K dataset. }
	\label{expadb}
\end{figure*}

\begin{table*}[!t]
  \centering
  \caption{Quantitative results on the Adobe Composition-1K dataset.  $\dag$ denotes that results are calculated on the whole image. 
  }
    \begin{tabular}{l|c|cccc}
        \toprule
    Method & {Trimap}&{SAD} & {MSE} & {Grad} & {Conn} \\
    \midrule
DIM~\cite{xu2017deep} &\checkmark&50.40  & 17.00  & 36.70  & 55.30  \\
IndexNet~\cite{lu2019indices}&\checkmark& 45.80  & 13.00  & 25.90  & 43.70   \\
GCAMatting~\cite{li2020natural} &\checkmark& 35.28  & 9.00  & 16.90  & 32.50  \\
PIIAMatting~\cite{wang2021ImprovingDeepImageMatting}&\checkmark& 36.40& 9.00 &16.90 &31.50\\
A2U~\cite{dai2021learning} &\checkmark& 32.10  & 7.80  & 16.33  & 29.00  \\
HDMatt~\cite{yu2020high}                &\checkmark & 33.50  & 7.30   & 14.50  & 29.90 \\
TIMI-Net~\cite{Liu_2021_ICCV}&\checkmark& 29.08  & 6.00  & 11.50  & 25.36   \\
SIM~\cite{sun2021sim}  &\checkmark&27.70  & 5.60  & 10.70  & 24.40  \\
FBAMatting~\cite{forte2020fbamatting}&\checkmark & 26.40  & 5.40  & 10.60  & 21.50  \\
LSAMatting~\cite{lsam} &\checkmark& 25.90  & 5.40  & 9.25  & 21.50  \\
TransMatting~\cite{cai2022TransMatting}&\checkmark& 24.96 &4.58& 9.72& 20.16  \\
MatteFormer~\cite{park2022matteformer}&\checkmark& 23.80  & 4.03  & 8.68  &18.90  \\
DCAM (Ours)                            &\checkmark& \bf{22.62}  &\bf{3.34}   & \bf{7.67}   & \bf{18.02} \\
    \midrule
DIM$^\dag$~\cite{Zhang2019A} &\checkmark& 52.76& 6.36&32.16& 53.03\\
IndexNet$^\dag$~\cite{lu2019indices}&\checkmark & 45.30& 5.12& 26.01 &43.16\\
GCAMatting$^\dag$~\cite{li2020natural}                         &\checkmark  &35.82 &{3.69}& 18.66 &33.14 \\ 
Late Fusion$^\dag$~\cite{Zhang2019A} &×& 58.34& 11.00& 41.63 &59.74\\
HAttMatting$^\dag$~\cite{qiao2020attention}&× & 44.01 &7.00 &29.26 &46.41\\
    DCAM (Ours)$^\dag$                         &×  &  \bf{29.72} & \bf{2.09}   &  \bf{13.50}   &  \bf{26.11} \\ 
        \bottomrule
    \end{tabular}%
  \label{tab:adb}%
\end{table*}%

\noindent \textbf{Trimap-based Interactive Matting}: We compare DCAM with state-of-the-art trimap-based interactive matting methods, including DIM~\cite{xu2017deep}, IndexNet~\cite{lu2019indices}, GCAMatting~\cite{li2020natural}, A2U~\cite{dai2021learning}, PIIAMatting~\cite{wang2021ImprovingDeepImageMatting}, HDMatt~\cite{yu2020high}, TIMI-Net~\cite{Liu_2021_ICCV}, FBAMatting~\cite{forte2020fbamatting}, LSAMatting~\cite{lsam}, TransMatting~\cite{cai2022TransMatting}, SIM~\cite{sun2021sim}, and MatteFormer~\cite{park2022matteformer}. 
All these methods are trained on the Adobe Composition-1K dataset. 
We summarize the qualitative and quantitative results in Table~\ref{tab:adb} and Figure~\ref{expadb}. 
The results in Table~\ref{tab:adb} indicate that DCAM outperforms all trimap-based interactive matting methods in all four evaluation metrics,  demonstrating its effectiveness.
Figure~\ref{expadb} shows that existing trimap-based  methods struggle to predict alpha mattes when the foreground and background colors are similar, while DCAM performs well in such scenarios.

\noindent \textbf{Automatic Matting}: We compare DCAM with several trimap-based interactive matting methods and automatic matting methods trained on the Adobe Composition-1K dataset, including DIM~\cite{xu2017deep}, IndexNet~\cite{lu2019indices}, GCAMatting~\cite{li2020natural}, LateFusion~\cite{Zhang2019A}, HAttMatting~\cite{qiao2020attention}. 
We present the quantitative results calculated on the whole image in Table~\ref{tab:adb}.
The table shows that existing automatic methods perform worse than trimap-based methods, indicating that automatic matting is more challenging.
However, our DCAM not only outperforms LateFusion, HAttMatting but also outperforms trimap-based methods such as DIM, IndexNet, and GCAMatting, demonstrating the high performance of DCAM.

\subsubsection{Results on Distinctions-646}
We conduct experiments on the Distinctions-646 dataset to evaluate the generalization ability of DCAM on trimap-based interactive matting and the performance of DCAM on automatic matting.

\begin{figure*}[!t]
	\centering{
			\includegraphics[width=1\linewidth]{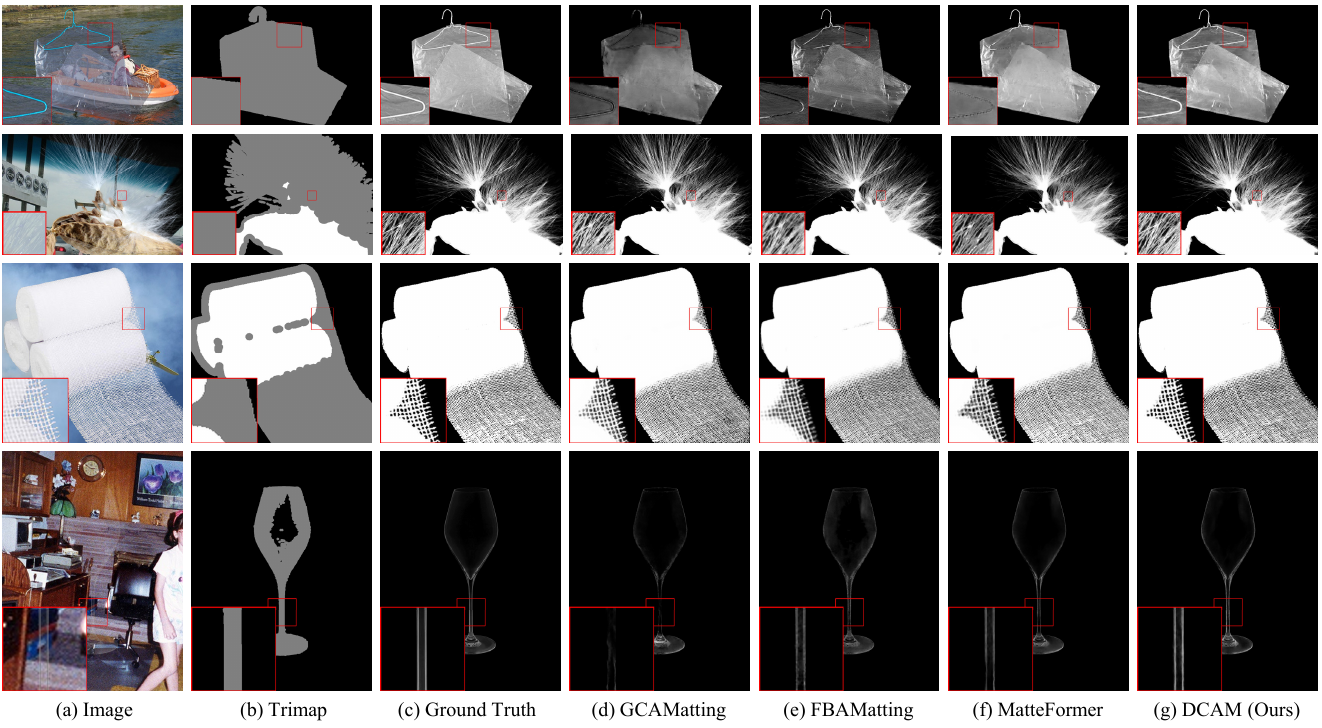}
	}
	\caption{Qualitative results on the Distinctions-646 dataset. All methods are trained on the Adobe Composition-1K dataset. }
	\label{exp646}
\end{figure*}

\noindent \textbf{Trimap-based Interactive Matting}: We compare our DCAM with trimap-based interactive matting methods, including DIM, IndexNet, GCAMatting, FBAMatting, TIMI-Net, and MatteFormer, which are all trained on the Adobe Composition-1K dataset. We summarize both quantitative  and qualitative results in Table~\ref{tab:d646} and Figure~\ref{exp646}, respectively. As shown in Table~\ref{tab:d646}, DCAM achieves superior performance on all four evaluation metrics. Moreover, as demonstrated in Figure~\ref{exp646}, DCAM outperforms previous methods in predicting accurate alpha mattes in large unknown regions, which are challenging for previous methods. These qualitative and quantitative results demonstrate the strong generalization ability of DCAM.

\begin{table*}[!t]
  \centering
  \caption{Quantitative results on the Distinctions-646. $\dag$ denotes the method is trained on the Adobe Composition-1K dataset. } 
    \begin{tabular}{l|c|cccc}
        \toprule
    Method & {Trimap}& {SAD} & {MSE} & {Grad} & {Conn} \\
    \midrule
    DIM$^\dag$~\cite{xu2017deep} &\checkmark&63.88 & 25.77 & 53.23 & 66.31 \\ 
    IndexNet$^\dag$~\cite{lu2019indices} &\checkmark& 44.93 &9.23 & 41.30 &44.86\\ 
    TIMI-Net$^\dag$~\cite{Liu_2021_ICCV}  &\checkmark& 42.61 & 7.75& 45.05 & 42.40 \\ 
    GCAMatting$^\dag$~\cite{li2020natural}  &\checkmark& 36.37 &8.19 & 32.34 &36.00\\
    FBAMatting$^\dag$~\cite{forte2020fbamatting} &\checkmark & 32.28 &5.66 & 25.52 & 32.39 \\
    MatteFormer$^\dag$~\cite{park2022matteformer} &\checkmark & 23.60 & 3.12 & 13.56 & 21.56 \\
    DCAM (Ours)$^\dag$   &\checkmark &\bf{20.72}&\bf{2.83}&\bf{11.32}&\bf{19.12}   \\   
    \midrule
    HAttMatting~\cite{qiao2020attention} &×& 48.98 &9.40 &41.57& 49.93 \\
    HAttMatting++~\cite{hattplus} &×& 47.38& 8.80& 40.09& 45.60\\
    DCAM (Ours)   &× &\bf{31.27}&\bf{4.86}&\bf{25.50}&\bf{31.72}   \\   
    \bottomrule
    \end{tabular}%
  \label{tab:d646}%
\end{table*}%

\noindent \textbf{Automatic Matting}: We also evaluate DCAM against automatic matting methods, including HAttMatting and HAttMatting++\citep{hattplus}, which are trained on the Distinctions-646 dataset. 
The quantitative results are summarized in Table~\ref{tab:d646}. 
Automatic matting methods perform worse than trimap-based methods, highlighting the difficulty of automatic matting. However, our DCAM outperforms HAttMatting, HAttMatting++, and many trimap-based methods such as DIM, IndexNet, and GCAMatting, demonstrating its superior performance.

\subsubsection{Results on P3M}
To evaluate the performance of DCAM on automatic image matting, we compare it with automatic matting methods on the P3M dataset. Specifically, we evaluate MODNet, P3M, GFM, SHM, MatteFormer, and our DCAM, which are trained on P3M, on two test sets of P3M: P3M-500-NP and P3M-500-P. We use SAD, MAD, and MSE as evaluation metrics, and summarize the quantitative results in Table~\ref{table_p3ma}.
As indicated in Table~\ref{table_p3ma}, DCAM outperforms all existing automatic image matting methods, which demonstrates its superior performance.

\begin{table}[!t]
\caption{Quantitative comparison on the P3M-500-NP and P3M-500-P test sets. }
\label{table_p3ma}
\begin{tabular*}{\textwidth}{@{\extracolsep\fill}l|ccc|ccc}
\toprule%
& \multicolumn{3}{@{}c|@{}}{P3M-500-NP} & \multicolumn{3}{@{}c@{}}{P3M-500-P} \\\cmidrule{2-4}\cmidrule{5-7}%
Model & SAD & MSE & MAD &  SAD & MSE & MAD  \\
\midrule
	HAttMatting~\cite{qiao2020attention} &30.53&0.0072 &0.0176&25.99&0.0054 &0.0152\\
	SHM~\cite{chen2018semantic}&20.77  &0.0093 &0.0012&21.56  &0.0100 &0.0125\\
	LFM~\cite{Zhang2019A} & 32.59& 0.0131&0.0188& 42.95& 0.0191&0.0250\\
	GFM~\cite{li2022matting} &15.50&0.0056&0.0091&13.20&0.0050&0.0080\\
 P3M~\cite{2021Privacy} &11.23&0.0035&0.0065&8.73&0.0026&0.0051\\
 MatteFormer~\cite{park2022matteformer} &{11.43}&{0.0034}&{0.0066}&{9.46} &{0.0028}&{0.0055}\\
	\midrule
	DCAM (Ours)  &\bf{8.92}&\bf{0.0029}&\bf{0.0053}&\bf{7.45} &\bf{0.0022}&\bf{0.0044}\\
\bottomrule
\end{tabular*}
\end{table}

\begin{table*}[!t]
	\caption{Quantitative comparison on the PPM-100 dataset.  $\dag$ denotes that the click guidance is adopted. }
	\label{table_ppm}
	\centering
	\begin{tabular}{l|cc}
    \toprule
	Method  & MSE & MAD \\
    \midrule
	DIM~\cite{xu2017deep} &0.0115 & 0.0178\\
	FDMPA~\cite{zhu2017fast}  &0.0101 &0.0160\\
	HAttMatting~\cite{qiao2020attention} &0.0067 &0.0137\\
	SHM~\cite{chen2018semantic}  &0.0072 &0.0152\\
	LFM~\cite{Zhang2019A} &  0.0094&0.0158\\
	MODNet~\cite{ke2020is} &0.0044&0.0086\\
 MatteFormer$^\dag$~\cite{park2022matteformer}   &{0.0092}&{0.0151}\\
	\midrule
	DCAM (Ours)$^\dag$   &\bf{0.0035}&\bf{0.0065}\\
    \bottomrule
	\end{tabular}
\end{table*}

\begin{table*}[!t]
  \centering
  \caption{Comparison of the computational complexity and parameter amounts of image matting methods. }
    \begin{tabular}{l|c|c}
    \toprule
    Method & MACs (G) & Params (M) \\
    \midrule
    DIM~\cite{xu2017deep}   & 727.4 & 130.5  \\
    GCAMatting~\cite{li2020natural} & 257.3 & 24.1  \\
    FBAMatting~\cite{forte2020fbamatting} & 686.0   & 34.8  \\
    SIM~\cite{sun2021sim}   & 1001.9 & 44.5  \\
    TIMI-Net~\cite{Liu_2021_ICCV}  & 351.3 & 35.0  \\
    MatteFormer~\cite{park2022matteformer} & 233.3 & 44.9  \\
    \midrule
    DCAM (Ours) & 365.6 & 45.6  \\
    \bottomrule
    \end{tabular}%
  \label{tab:cs}%
\end{table*}%

\subsubsection{Results on PPM-100}
To evaluate the generalization ability of DCAM, we compare it with DIM, FDMPA~\cite{zhu2017fast}, LateFusion, HAttMatting, MatteFormer, and MODNet. In particular,  MatteFormer and DCAM are trained on HIM-100K, while DIM, FDMPA, LateFusion, HAttMatting, and MODNet are trained on the private training set of PPM-100. Furthermore, MatteFormer and DCAM adopt the click guidance. We follow MODNet to use MAD and MSE as evaluation metrics, and summarize the quantitative results in Table~\ref{table_ppm}. 
The results show that our DCAM surpasses existing automatic matting methods, indicating its capability to accurately estimate the alpha mattes for real-world portrait images.

\subsection{Model Complexity}
To assess the model complexity of the proposed DCAM, we compare the  computational complexity and parameter amounts of DCAM with trimap-based interactive matting methods including DIM, GCAMtting, FBAMatting, SIM, and MatteFormer. 
The computational complexity  of the model refers to the number of multiply–accumulate operations (MACs) required to perform inference on a $1024 \times 1024$ image. 
The results are summarized in Table~\ref{tab:cs}. 
As shown in Table~\ref{tab:cs}, DCAM does  exhibit slightly higher computational complexity and parameter amounts compared to MatteFormer, GCAMatting, and TIMI-Net.
This discrepancy can be attributed to the innovative design of DCAM, which incorporates the Global Object Aggregator and Local Appearance Aggregator within an encoder-decoder network to aggregate global and local contexts, thereby achieving robust performance across various matting tasks.
Although the introduction of these network structures enhance the universality, it also slightly increases the model complexity. 
Nevertheless, the model complexity of DCAM  remains on par with mainstream methods like FBAMatting and SIM. This highlights that the performance of DCAM is not achieved through an increase in model complexity.


\begin{table*}[!t]
  \centering
  \caption{Ablation study on the DCAM framework. NCA indicates the number of cascading aggregators. GEM indicates the guidance embedding layer. OBJ and OBJ-SEM indicate the global object aggregator adopts the object features and object-semantics features, respectively. TRAN and HTRA indicate the local appearance aggregator adopts the Transformer structure and the hybrid Transformer structure, respectively.}
  \resizebox{1\linewidth}{!}{
    \begin{tabular}{c|cccccc|cc}
    \toprule
    Model &NCA & GEM & OBJ&OBJ-SEM &TRAN &HTRA & MSE & MAD  \\
    \midrule
  B1 &   0     &    -   &   -    &    -   & -      &  -     & 0.004136 & 0.005532 \\ 
    
  B2 &   1     &    -   & -     & \checkmark     & -    & \checkmark     & 0.002578 & 0.003867 \\
    
  B3 &   1     & \checkmark     & -    & \checkmark     & -    & \checkmark     & 0.002441 & 0.003738 \\
    
  B4 &   2     & \checkmark     & \checkmark     &    -   &   -    &   -    & 0.002775 & 0.004076 \\
    
  B5 &   2     & \checkmark     & -     & \checkmark     &    -   &    -   & 0.002535 & 0.003841 \\
    
  B6 &   2     & \checkmark     &  -     &    -   & \checkmark     &    -   & 0.002851 & 0.004172 \\
    
  B7 &   2     & \checkmark     &  -     &    -   & -    & \checkmark     & 0.002618 & 0.003929 \\
    
  B8 &   2     & \checkmark     & -    & \checkmark     & -    & \checkmark     & \textbf{0.002280}  & \textbf{ 0.003422} \\ 
    \bottomrule 
    \end{tabular}%
    }
  \label{tab:as}%
\end{table*}%

\subsection{Ablation Study}
To evaluate the effectiveness of  the proposed improvements in DCAM, we conduct diagnostic experiments on HIM-100K and summarize the results in Table~\ref{tab:as}.

\noindent \textbf{Guidance  Embedding Layer.}
We adopt a guidance embedding layer to enhance the guidance information in the context features. To validate the effectiveness of this design, we evaluate the DCAM model with or without the guidance embedding layer. As shown in Table~\ref{tab:as}, the results of models \textbf{B2} and \textbf{B3} demonstrate the DCAM model with the guidance embedding layer outperforms the one without it, which verifies the effectiveness of the guidance embedding layer.

\noindent \textbf{Global Object Aggregator.}
We introduce global object aggregators to globally aggregate context features. To verify the effectiveness of the global object aggregator and to test the advantages of the object-semantics features over object features, we conduct experiments on the basic network with three different settings: (1) without any global object aggregator, (2) with an object feature-based global object aggregator, and (3) with an object-semantics feature-based global object aggregator. As shown in Table~\ref{tab:as}, the results of models \textbf{B1}, \textbf{B4}, and \textbf{B5} show that the basic network with object-semantics feature-based global object aggregator outperforms the others, demonstrating the effectiveness of global object aggregators and the superiority of object-semantics features over object features.

\noindent \textbf{Local Appearance Aggregator.}
We introduce local appearance aggregators  for local context aggregation. To verify the effectiveness of the local appearance aggregator and the hybrid Transformer structure, we conduct experiments on the basic network with three different settings: (1) without the local appearance aggregator, (2) with a Transformer-based local appearance aggregator, and (3) with a hybrid Transformer-based local appearance aggregator.
The results presented in Table~\ref{tab:as}, specifically models \textbf{B1}, \textbf{B6}, and \textbf{B7} show that the basic network with the hybrid Transformer-based local appearance aggregator performs the best, which  verifies the effectiveness of the local appearance aggregator and highlights the advantage of the hybrid Transformer.

\noindent \textbf{Cascading Aggregator.}
We adopt the cascading aggregator design to iteratively refine context features with multiple global object aggregators and local appearance aggregators. To validate the effectiveness of this design, we compare the DCAM models with 0, 1, and 2 global object aggregators and local appearance aggregators. 
The results of models \textbf{B1}, \textbf{B3}, and \textbf{B8} in Table~\ref{tab:as} demonstrate that the cascading approach achieves the best performance compared to the other settings, highlighting the effectiveness of the cascading design.

\section{Conclusion}
In this paper, we propose  a simple and universal matting framework, named Dual-Context Aggregation Matting (DCAM), which can perform robust image matting with arbitrary guidance or without guidance.
Specifically, DCAM utilizes a semantic backbone network to extract low-level features and context features from the input image and the guidance. 
Then, we introduce a dual-context aggregation network that incorporates global object aggregators and local appearance aggregators to iteratively refine the extracted context features.
Finally, we adopt a matting decoder network to fuse the low-level features and refine context features for alpha matte estimation. 
Extensive experimental results on five image matting datasets demonstrate that DCAM outperforms state-of-the-art matting methods in the automatic and interactive matting tasks, which  indicates its strong universality and high performance.
In the future, we will extend DCAM to enable a single network to perform automatic matting and interactive matting without requiring multiple training procedures.

\bibliography{references}

\end{document}